\documentclass[runningheads]{llncs}
\usepackage{epsfig,latexsym,soul,hyperref,xcolor,amsmath,amssymb,pifont,mathabx,dsfont}

\usepackage{subfigure}

\usepackage{graphicx}
\usepackage{bbm}
\begin{document}
\title{To Switch or not to Switch: \\ Predicting the Benefit of Switching between Algorithms based on Trajectory Features}
\titlerunning{To Switch or not to Switch}

%
%
\author{Diederick Vermetten\inst{1}\orcidID{0000-0003-3040-7162},
Hao Wang\inst{1}\orcidID{0000-0002-4933-5181},
Kevin Sim\inst{2}\orcidID{0000-0001-6555-7721},
Emma Hart\inst{2}\orcidID{0000-0002-5405-4413}}
\authorrunning{Vermetten et al.}
%
\institute{LIACS, Leiden University, Niels Bohrweg 1, NL-2333 Leiden, Netherlands\\
\email{\{d.l.vermetten,h.wang\}@liacs.leidenuniv.nl} \and
School of Computing, Edinburgh Napier University, Scotland, UK, 
\email{\{k.sim,e.hart\}@napier.ac.uk}}
\maketitle            
\begin{abstract}
Dynamic algorithm selection aims to exploit the complementarity of multiple optimization algorithms by switching between them during the search. While these kinds of dynamic algorithms have been shown to have potential to outperform their component algorithms, it is still unclear how this potential can best be realized. One promising approach is to make use of landscape features to enable a per-run trajectory-based switch. Here, the samples seen by the first algorithm are used to create a set of features which describe the landscape from the perspective of the algorithm. These features are then used to predict what algorithm to switch to. 

In this work, we extend this per-run trajectory-based approach to consider a wide variety of potential points at which to perform the switch. We show that using a sliding window to capture the local landscape features contains information which can be used to predict whether a switch at that point would be beneficial to future performance. By analyzing the resulting models, we identify what features are most important to these predictions. Finally, by evaluating the importance of features and comparing these values between multiple algorithms, we show clear differences in the way the second algorithm interacts with the local landscape features found before the switch. 

\keywords{Dynamic algorithm selection \and benchmarking \and exploratory landscape analysis.}
\end{abstract}

\section{Introduction}

Over the years, the field of optimization has developed a large variety of algorithms to tackle the wide range of optimization problems. While this growing diversity of solvers has obvious benefits, it also highlights some critical challenges, such as the issue of finding an appropriate algorithm to solve a particular optimization problem. 

This \textbf{algorithm selection} problem has been a topic of study for a long time~\cite{rice1976algorithm,DBLP:journals/ec/KerschkeHNT19}. Many approaches rely on a pre-computed set of features about the problem, most commonly based on exploratory landscape analysis (ELA)~\cite{mersmann2011exploratory}. This approach usually requires the collection of function evaluations before the actual exploration process is started, which could be an inefficient use of resources. In contrast, approaches which make use only of meta-information of the problem to be optimized, e.g. the dimension, variable types etc. have shown considerable success~\cite{DBLP:journals/tec/MeunierRWRRTMD22}, but they are inherently less granular in their selection.

When faced with an optimization problem, we would ideally be able to make use of the meta-information to select an initial algorithm configuration to start the search, and then switch to other algorithm if the search behaviour provides evidence that this would be beneficial. This approach is considered as \textbf{dynamic algorithm selection (dynAS)}, and might be able to exploit the benefits of ELA-based features without the need for pre-computing the features. Previous work has shown that while the ELA features collected during the search might differ from those collected beforehand, they still contain enough information to distinguish between different search landscapes~\cite{DBLP:conf/evoW/JankovicED21}. 

While previous work into dynAS has shown that switching between two algorithms during the optimization process has significant potential~\cite{gecco_towards_bbob_dynas}, there are still several obstacles which need to be overcome to create practical implementations. One major factor lies in the switching procedure itself: when starting an algorithm in the middle of a search, we need to warmstart it using the information collected up to that point. While some basic warmstarting mechanisms show promising results, there is still a lot of room for further improvement in this area~\cite{dominik_paper}.

In this paper, we analyze another key aspect of dynAS: the point at which the switch is initiated. Previous work has mostly focused on showing the potential performance gains of the overall dynAS approach, and make use of a fixed-target switching point to illustrate this~\cite{gecco_towards_bbob_dynas}. However, when using a specific function value or precision to the global optimum to determine when to switch, the black-box assumption is broken. While fixed-budget approaches can mitigate this by switching after a specific number of evaluations, these approaches are still impacted by the stochasticity of the algorithm, which can lead to poor performance and limited generalizability. 

Since the determination of an overall switching point on a per-function basis is impacted heavily by the specifics of the current search process at that time, we will instead consider a per-run, feature-based approach, where we predict the benefits of performing a switch at different points in the search process. This decision will be made based on the search history so far. We do this by building random forest regressors which take the local landscape features as input and predict whether switching at that point would be beneficial. By analyzing these models for different combinations of algorithms on a wide set of functions, we show that the local trajectory features contain enough information to gauge the potential benefits of switching, which could in future form the basis for a fully online policy for switching between algorithms. By analyzing the resulting models, we can additionally gain insight into the importance of the used features.  

\section{Background}

\subsection{BBOB}
In order to collect algorithm performance data, we make use of a set of noiseless, single-objective black-box optimization problems from the BBOB suite, as defined in the COCO platform~\cite{hansen2020coco}. This suite consists of 24 problems, which are widely used in the continuous optimization domain for performing benchmarking of iterative optimization heuristics. Each of these problems can be scaled to arbitrary dimensionality, and can be combined with transformations in both search- and objective space to create a set of problem instances which are said to preserve the global function properties of the original functions~\cite{bbobfunctions}. In order to access the problems, we make use of the IOHexperimenter~\cite{iohexperimenter} framework.

\subsection{ELA}
In order to determine when a switch is beneficial to the performance of the search, we make use of the current state of the optimization process. Previous work into dynamic algorithm selection~\cite{ppsn_per_run} has shown that using ELA features~\cite{mersmann2011exploratory} provides promising results. While ELA was originally defined as a way to capture the global properties of a function, using it on a part of the landscape as seen by an optimization algorithm has been shown to capture some local features~\cite{jankovic2019adaptive} which can be used to make decisions about the remainder of the search process. 

In particular, we make use of the flacco~\cite{flacco} library to compute the landscape features used in this paper. We focus only on the set of `cheap' features, since those do not require additional samples to calculate. In total, we consider 68 features, which have previously been used to distinguish between BBOB problems or to select problems with similar characteristics~\cite{DBLP:conf/evoW/RenauDDD21,paper_fuxing_gecco}.

\subsection{DynAS}

The notion that complementarity between the characteristics of different algorithm or algorithm variants / configurations can be exploited has been explored in areas like hybrid / memetic algorithm design~\cite{neri2012memetic}, but is also a key part of parameter control~\cite{karafotias2014parameter}. Dynamic algorithm selection and dynamic algorithm configuration~\cite{DBLP:conf/ecai/BiedenkappBEHL20,adriaensen2022automated} similarly aim to build upon the complementarity between algorithms by allowing us to switch from one algorithm to another during the search process. In order to avoid a loss of information, this second algorithm should be warm-started using the information gathered in the initial part of the search. In the case of DAC, this warmstarting can be straightforward when the base algorithm structure is fixed, e.g. when adapting strategy parameters of CMA-ES~\cite{DBLP:conf/ppsn/ShalaBAALH20,research_project}. Because of this, most  approaches for DAC are based on reinforcement learning~\cite{eimer2021dacbench}. However, the problem of DAC, and dynAS more specifically, can also be viewed as a hyperparameter optimization problem. 

While the potential of this viewpoint towards dynamic algorithm selection seems to be significant~\cite{gecco_towards_bbob_dynas}, there are many open challenges which need to be addressed to realize these benefits. In particular, modifying the point at which the switch is performed will greatly change the relative importance of the other parameters, which is challenging for most hyperparameter tuning methods to deal with. As such, approaches which do not rely purely on static tuning, but rather use tuning to find models which can be used online, e.g. through finding relations between local landscape features and per-run optimal switching decisions, seems to be promising~\cite{ppsn_per_run}. In this work, we extend this per-run trajectory-based switching by incorporating the variability of the switching point. 

\section{Experimental Setup}\label{sec:setup1}

\textbf{Reproducibility} The code and data used in this paper has been made available on Zenodo~\cite{repository_reproducibility}. A description of the steps required to recreate the results is included in this repository. For each of the figures shown for a particular function / algorithm / setting, equivalent figures for the remaining configurations are made available on Figshare~\cite{repository_reproducibility}, in addition to several figures which could not be included due to space constraints in this paper.

\subsection{Algorithm Portfolio}
Since the potential of switching between algorithms seems to be highly dependent on the set of algorithms considered in the used portfolio~\cite{gecco_towards_bbob_dynas}, we consider a set of 5 algorithms:
\begin{itemize}
    \item Covariance Matrix Adaptation Evolution Strategy \textit{CMA-ES}~\cite{hansen_adapting_1996} (implementation from the modcma package~\cite{nobel_modcma_assessing})
    \item Differential Evolution \textit{DE}~\cite{price1997differential} (implementation from nevergrad~\cite{nevergrad})
    \item Particle Swarm Optimization \textit{PSO}~\cite{kennedy1995particle} (implementation from nevergrad)
    \item Success-History based Adaptive
Differential Evolution \textit{SHADE}~\cite{shade} (implementation from pyade~\cite{pyade})
    \item Constrained Optimization By Linear Approximation \textit{Rcobyla}~\cite{powell1994direct} (implementation from nevergrad)
\end{itemize}

\begin{figure}
    \centering
    \includegraphics[width=\textwidth]{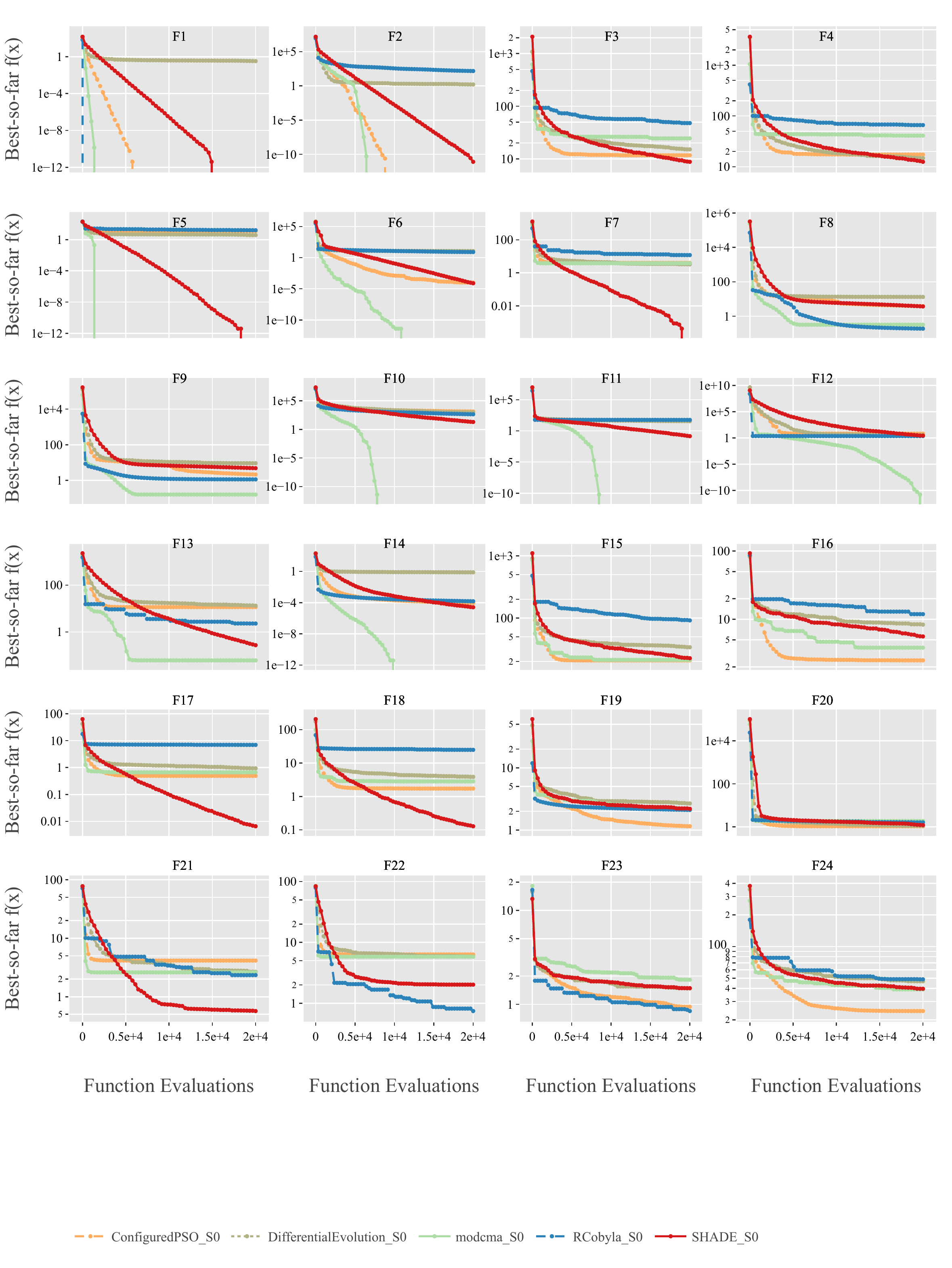}
    \caption{Mean function value reached relative to the used budget, for 24 BBOB functions. Figure generated using IOHanalyzer~\cite{IOHanalyzer}. Data available for interactive visualization at \url{iohanalyzer.liacs.nl} (dataset source 'DynAS\_EvoStar23').}
    \label{fig:perf_overall}
\end{figure}

\begin{figure}
    \centering
    \includegraphics[width=\textwidth]{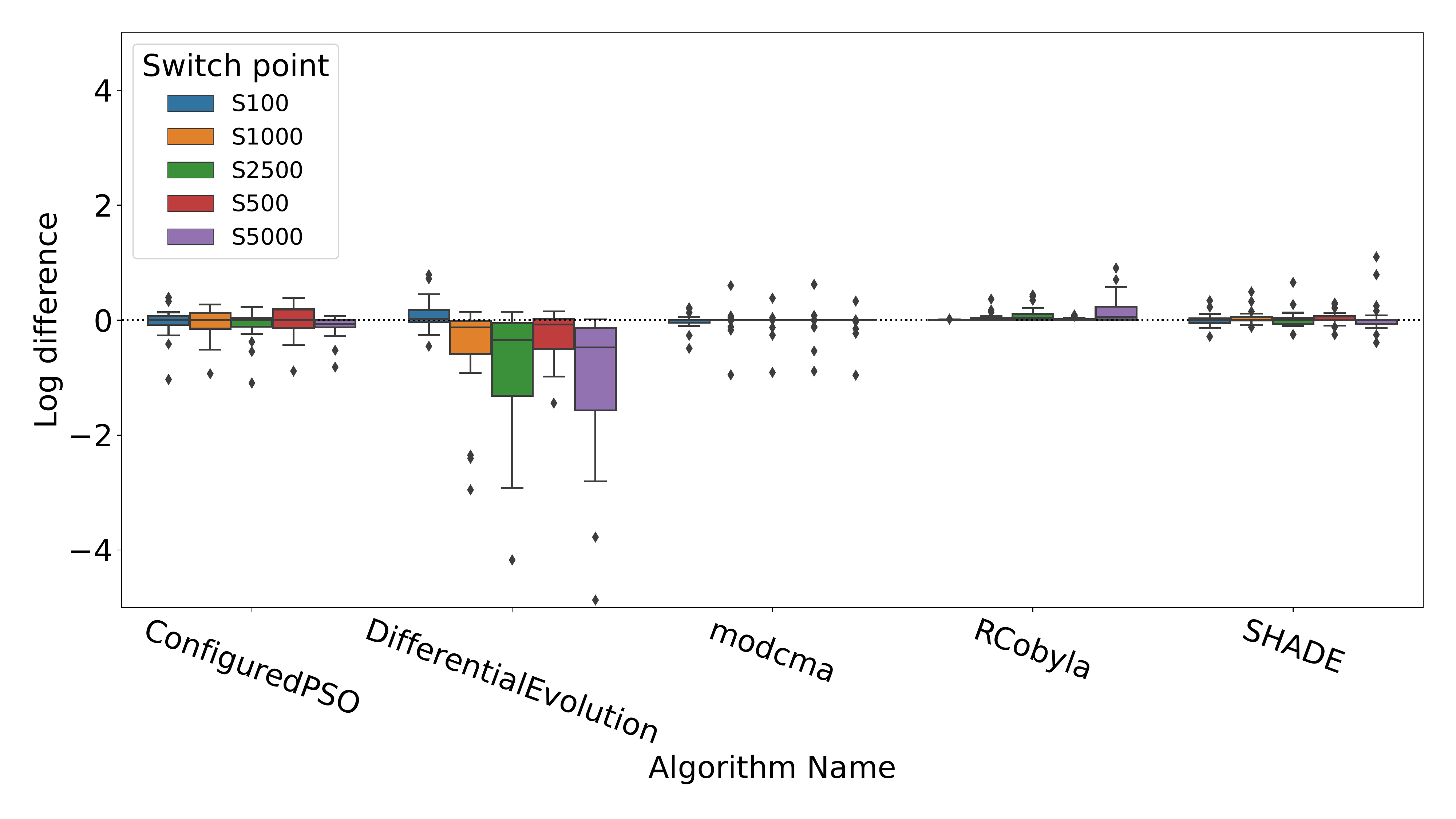}
    \caption{The log-distance between geometric mean of function value reached after $10\,000$ evaluations (limited to $10^{-8}$). Differences are computed as mean with restart minus mean without restart, so negative values indicate restarts improve performance. Each box represents 24 10-dimensional BBOB problems, for each of the 5 algorithms in the portfolio for the set of 5 tested switching points. }
    \label{fig:warmstart_impact}
\end{figure}

We show the performance of these 5 algorithms on the 10-dimensional BBOB problems from the fixed-budget perspective in Figure~\ref{fig:perf_overall}. We see that there are significant differences in the performances of these algorithms, with no algorithm consistently dominating all others.

In addition to the algorithms, we implement a warm-starting mechanisms to be able to switch between them. For the nevergrad-based algorithms, we make use of the built-in \textit{ask-not-told} functionality, which adapts the state of the algorithm based on a set of observations ($\{x, f(x)\}$). For starting the CMA-ES we use the warmstarting mechanism proposed in~\cite{dominik_paper}, which sets the center of mass and stepsize based on the $3$ best solutions found so far. For switching to SHADE, we initialize the population as the last N points seen by the previous algorithm, where $N$ is the population size.

To illustrate the usability of these warmstarting mechanisms, we investigate the performance achieved by switching from each algorithm to itself, using the described warm-starting mechanism. Since each of these warmstarting mechanisms inherently loses some information about the search process, we assume the warm-started versions will have slightly worse performance than their equivalent non-warmstarted runs. The results of running each of the 5 algorithms with 5 different points at which they are warm-started, are visualized in Figure~\ref{fig:warmstart_impact}. From this figure, we see that the performance loss from warm-starting is relatively minor, indicating that most of the relevant information is passed to the second part of the search. 

\subsection{Finding usecases using irace}\label{sec:irace}

\begin{figure}
    \centering
    \includegraphics[width=\textwidth]{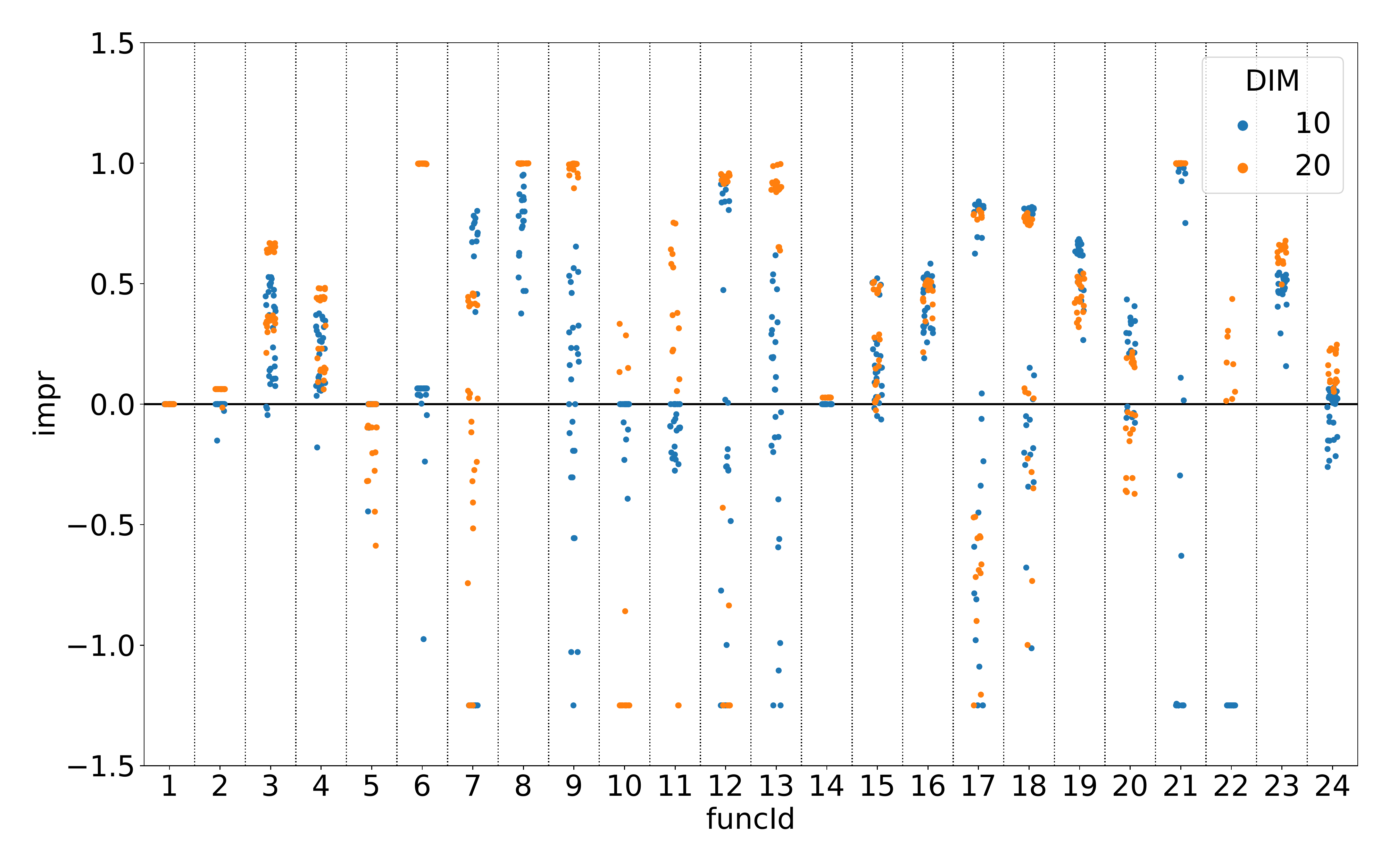}
    \caption{Relative improvement in terms of geometric mean of final function value of the elite configurations of irace against the virtual best solver (best static algorithm per function/dimension). Negative improvements are capped at $-1.25$ for visibility.  }
    \label{fig:rel_impr_means}
\end{figure}

\begin{figure}
    \centering
    \includegraphics[width=\textwidth,trim=9mm 11mm 4mm 5mm,clip]{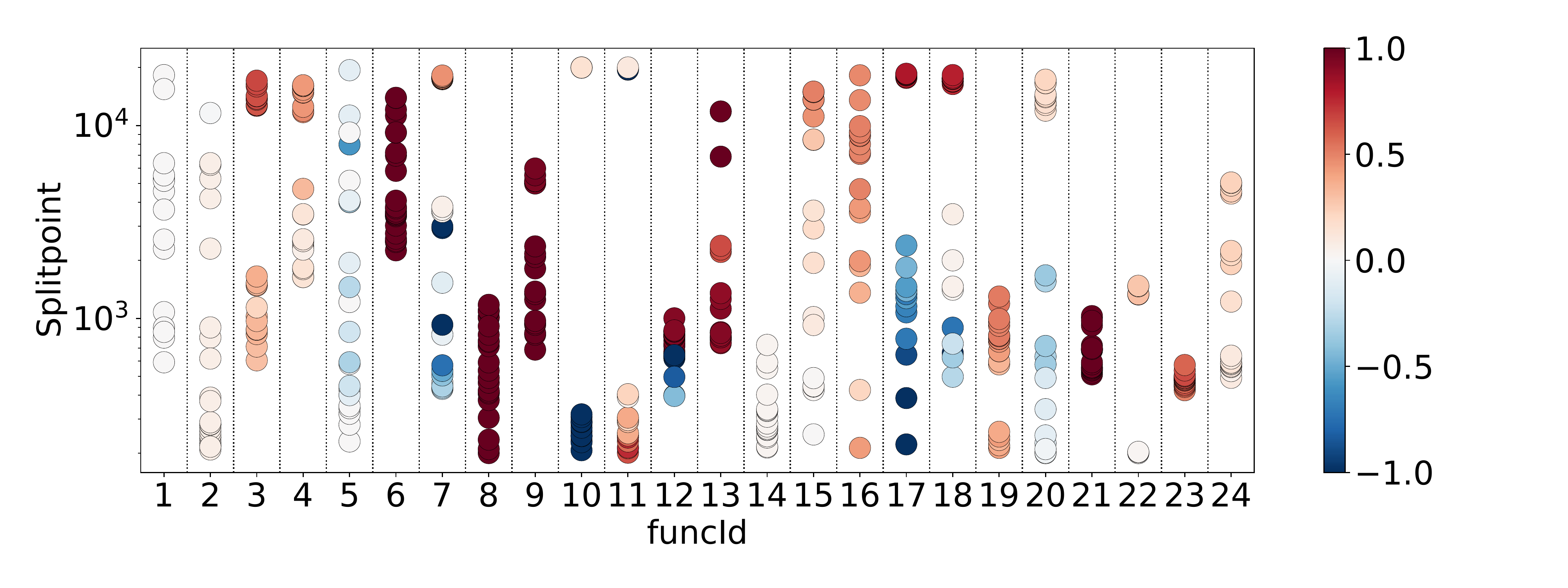}
    \caption{Distribution of the switch point in the elite configurations found by irace, for the 20-dimensional versions of the BBOB functions. The color of the dots corresponds to the relative improvement over VBS as shown in Figure~\ref{fig:rel_impr_means} }
    \label{fig:splitpoint_to_impr}
\end{figure}

To identify whether the selected portfolio can benefit from dynamically switching between algorithms, we view the problem of dynamic algorithm selection from the perspective of hyperparameter tuning. We consider the dynamic algorithm to consist of three distinct parts: the first algorithm, the point at which to switch, and the second algorithm. We use irace~\cite{irace} to find the configurations which reach the best function value after $5\,000$ function evaluations. Since irace is inherently stochastic, we perform $5$ independent runs, and for each of the sets of elite configurations we perform $250$ verification runs ($50$ runs on $5$ instances). The performance of these configurations is then compared to the best static algorithm in the portfolio for each function (virtual best solver). This relative measure is visualized in Figure~\ref{fig:rel_impr_means}. 

From this figure, we can see that on most problems, there are sets of configurations which seem to outperform the static algorithms. However, for some cases we see deterioration in performance compared to the VBS, indicated by negative values. This can be explained partly by the stochasticity of the algorithms: the performance observed by irace is based on a limited number of runs, and by selecting based on these limited samples can be sub-optimal when looking at the true performance distribution~\cite{DBLP:conf/gecco/VermettenW0DB22}. Additionally, there might be some cost associated with the warmstarting when the samples are collected from an initial algorithm which is not the same as the algorithm being switched to. 

Since we see that there are some cases where a switch between algorithms appears beneficial, we can delve deeper into the configurations which show these benefits. In particular, we can look at the distribution of the used switch point and its correlation to the relative performance improvement, as is shown in Figure~\ref{fig:splitpoint_to_impr}. Here, we observe that the switch points are fairly widely distributed, and that multiple different switching points can lead to similar improvements in performance.

\begin{figure}
    \centering
    \includegraphics[width=\textwidth,trim=9mm 10mm 4mm 11mm,clip]{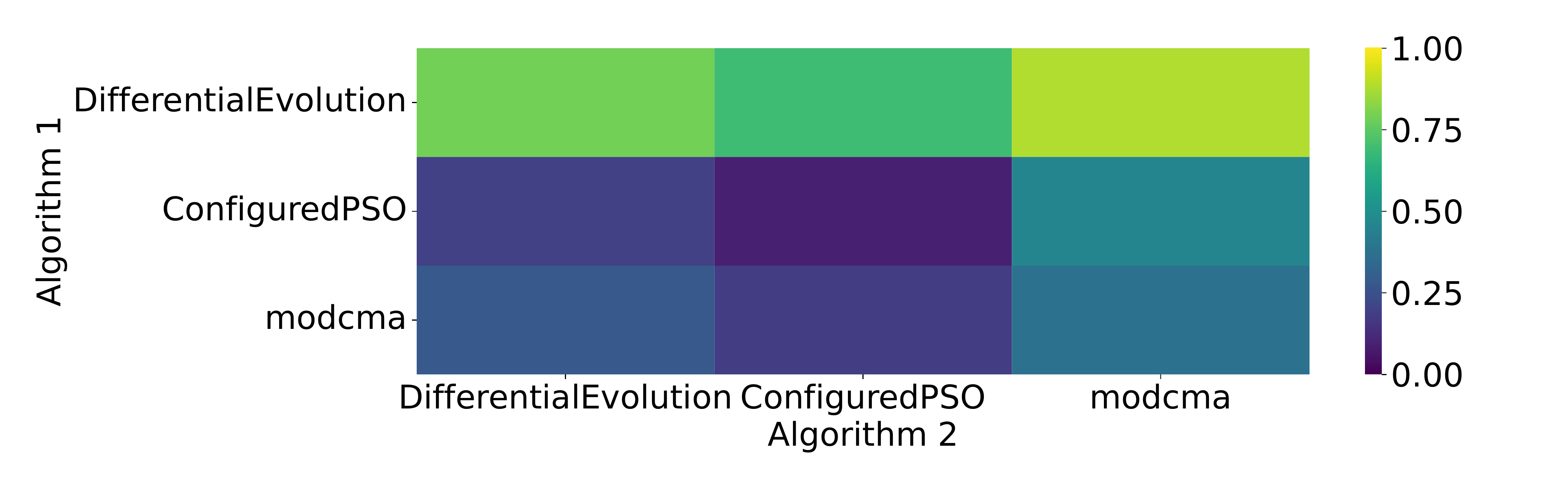}
    \caption{Fraction of cases in which a switch from algorithm 1 (y-axis) to algorithm 2 (x-axis) is beneficial.}
    \label{fig:overall_benefit_matrix}
\end{figure}

\section{Predicting Benefits of Switching}

While the setup as described in Section~\ref{sec:setup1} allows us to investigate the dependence of performance of a dynamic algorithm selection on the time at which the switch occurs, it does not provide directly usable insights into how this switch might be detected during the search. In order to investigate this online detection, we require a set of data where multiple switching points are attempted, such that we are able to identify on a per-run basis how beneficial each decision is. In addition, we collect features at each decision point, which can then be used to create a model to predict the observed benefits. 

\begin{figure}
    \centering
    \includegraphics[width=\textwidth]{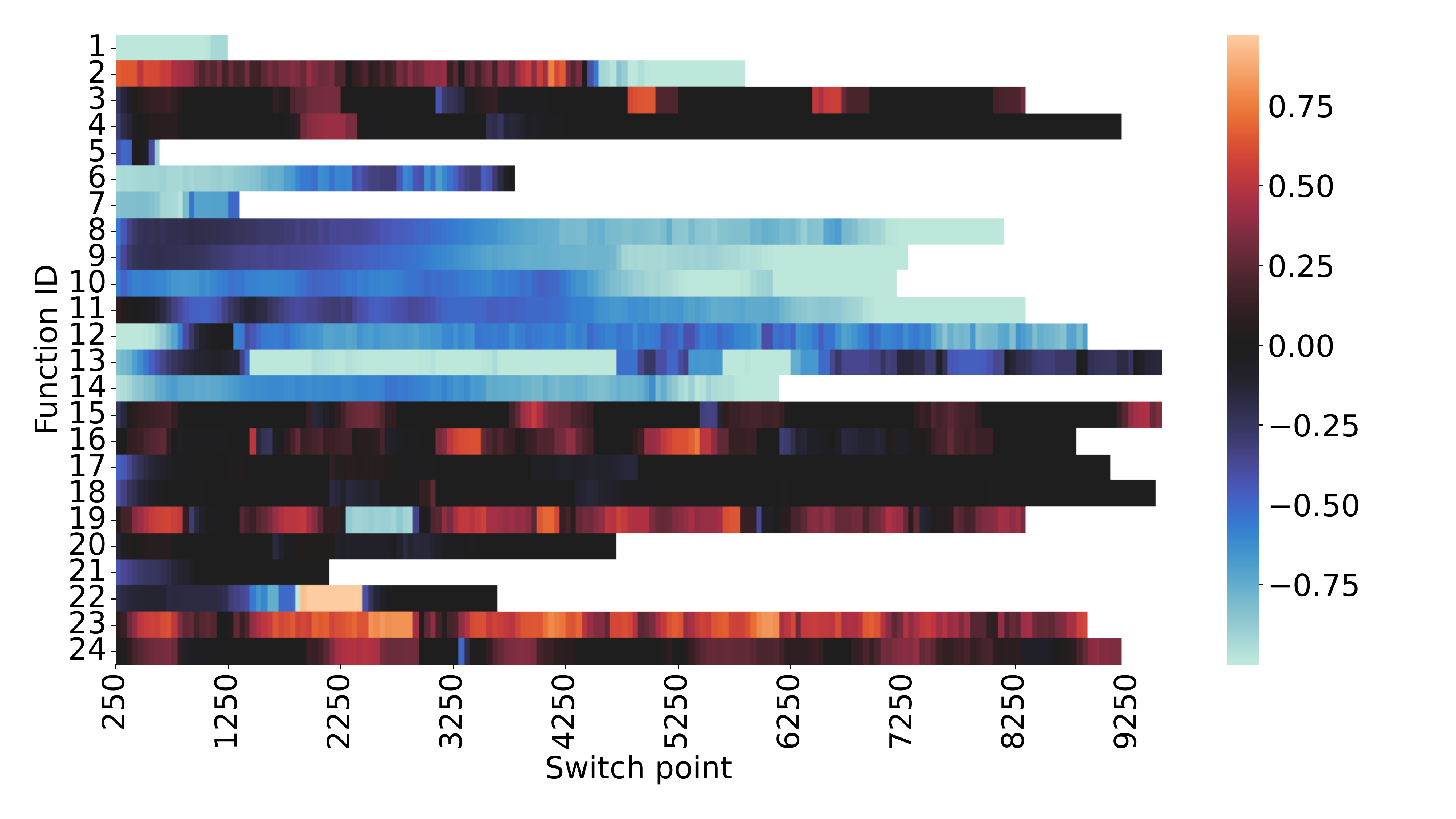}
    \caption{Mean relative benefit of switching from CMA-ES to DE at each of the selected switching points, for each of the 24 BBOB functions.}
    \label{fig:heatmap_overall_benefit}
\end{figure}

\subsection{Setup}
To achieve this, we set up a large-scale experiment collecting the performance data for a reduced portfolio of 3 algorithms (CMA-ES, PSO and DE) on all 24 10-dimensional BBOB problems. This reduction is done to reduce computation costs. We collect 5 runs on each of the first 5 instances, and collect the full trajectory of the static algorithm up to 10000 evaluations. Then, for all switch points linearly spaced from 50 to 9500, we collect the performance data achieved when switching to each of the 3 algorithms (so we include a switch to the selected algorithm to itself) in the portfolio after another 500 evaluations.We then consider the best fitness value reached in these 500 evaluations as the achieved performance of the dynamic algorithm. This short time-window is used to allow for the eventual creation of a dynamic switching regime which can perform more than one change during the optimization process. 

In Figure~\ref{fig:overall_benefit_matrix} we show the fraction of cases in which a switch provides benefit over continuing the first algorithm in these 500 evaluations. From this, we see that switching is often beneficial, particularly in the case of switching to CMA. This matches our observations from Section~\ref{sec:setup1}, where we saw that our chosen version of DE often benefits from restarts, while the CMA-ES is the best preforming algorithm overall.

To enable an easier comparison between the algorithms, we define the target value for our model to be the relative benefit of switching after $500$ evaluations, which is defined as follows:
\begin{equation}
    r(a_s, a_r) = \left(1-\frac{\texttt{min}(a_s,a_r)}{\texttt{max}(a_s,a_r)}\right) (2\cdot \mathbbm{1}_{a_s < a_r} - 1)
\end{equation} where $a_s$ is the performance when a switch is performed, and $a_r$ is the performance when no switch occurs.
This measure takes values in $[-1, 1]$, where positive values correspond to situations where switching is beneficial, while a negative value indicates detrimental effect of the switch. 

To highlight the overall importance of the switching point, we can visualize the mean relative benefit of switching at each point in a heatmap, as is done in Figure~\ref{fig:heatmap_overall_benefit} for the case of switching from CMA-ES to DE. Here, we see that even though the individual algorithm performance from Figure~\ref{fig:perf_overall} showed that CMA-ES dominates DE in most problems, and Figure~\ref{fig:overall_benefit_matrix} showed that this combination is not the most promising overall, there are still many cases where a switch would still be beneficial for the performance in the next $500$ evaluations. In particular, we see some clear distinctions between functions where switching is detrimental and some functions where benefits are observed, although not for all possible switching points.

In order to predict the benefit of switching at each decision point, we train a random forest model for each switch combination which outputs the relative benefit of performing the switch. The input for this model consists of the ELA features calculated on the trajectory of the first algorithm during the last $\{50,150,250\}$ evaluations before the switching point. We exclude the ELA features that require addition sample points, e.g., the so-called cell mapping features, resulting in 68 features in total.

This set is extended by including the diversity in the samples, both the mean component-wise standard deviation of the full set of samples (\textit{pop\_div}) and the standard deviation from their corresponding fitness values (\textit{fit\_div}). 

Features which are constant for all samples or give NaN values for more than $90\%$ of samples are removed from consideration. Features are then normalized (to zero mean and unit variance). 

The random forest models use the default hyperparameters from sklearn~\cite{scikit-learn}. Their performance is evaluated using the leave-one-function out strategy, where we train on the data from $23$ BBOB functions and use the remaining one for testing. This is repeated for each function, and the results shown in this section are always on this unseen function. For our accuracy measure, we make use of the mean square error.

\subsection{Results}

\begin{figure}
    \centering
    \includegraphics[width=\textwidth]{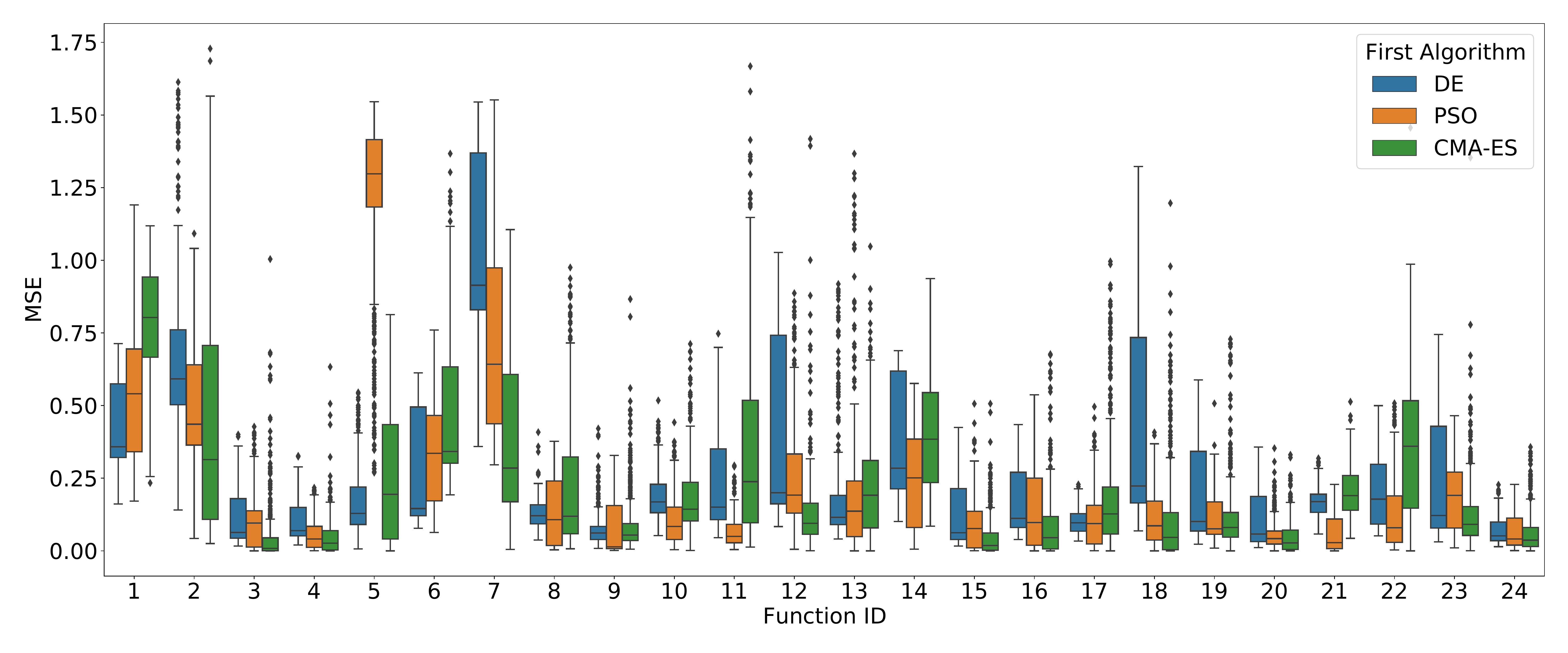}
    \caption{Distribution of model quality (Mean Square Error) for each function, colored according to the algorithm from which the switch occurs. Aggregated across the secondary algorithms and switch points.}
    \label{fig:mse_overall}
\end{figure}

\begin{figure}[t]
    \centering
    \begin{subfigure}
         \centering
         \includegraphics[width=0.98\textwidth, trim={0mm, 18.8mm, 0mm, 0mm},clip]{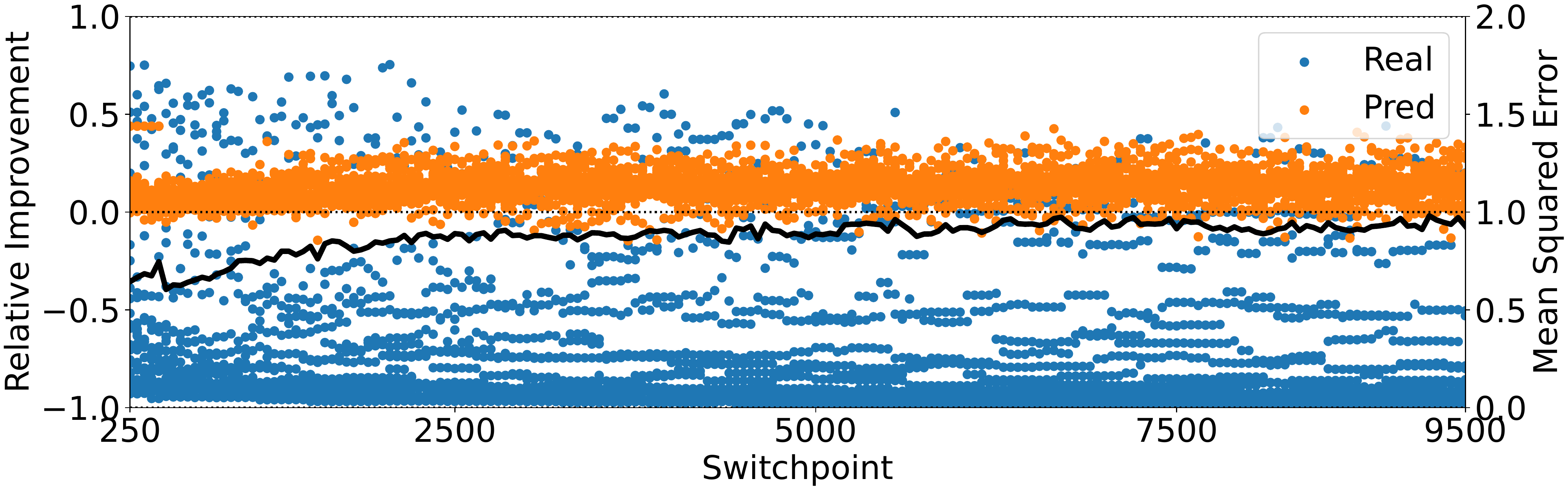}
          \label{F7_pred_vs_real_swpoint}
     \end{subfigure}
    \begin{subfigure}
         \centering
         \includegraphics[width=0.98\textwidth, trim={0mm, 0mm, 0mm, 0mm},clip]{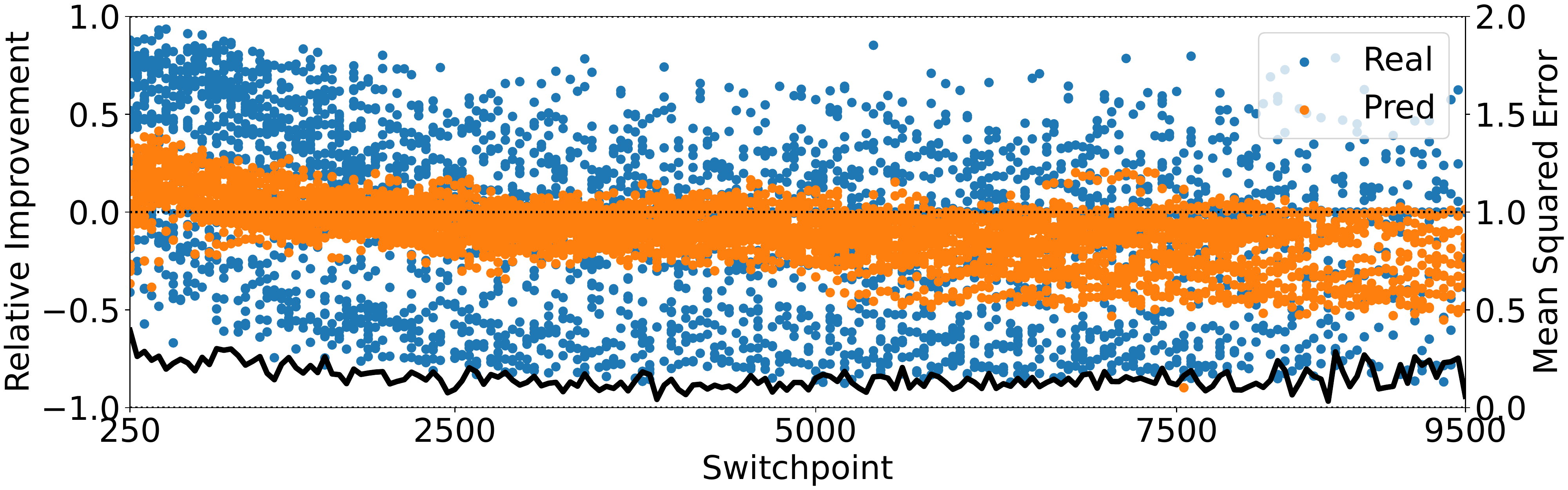}
         \label{F15_pred_vs_real_swpoint}
     \end{subfigure}
     
    \caption{Relation between improvement and point at which the switch occurs in, for both the real improvement and the improvement predicted by the RF model. Top: switching from DE to PSO on F7. Bottom: switching from PSO to CMA on F15. The thick black line shows the MSE of the model evaluated on the selected switch point only. X-axis is shared between the two subfigures.}
    \label{fig:pred_vs_real_swpoint}
\end{figure}

In Figure~\ref{fig:mse_overall}, we show the overall model quality per decision point, aggregated over the algorithm which is being switched to. This aggregation allows us to gain an overview of the potential to learn the relative benefit of switching from data, which illustrates significant differences among test functions and the choice of the first algorithm. From this figure, we can see that some settings lead to very poor MSE values. This can either indicate that the model is not able to extract the needed information from the training features, or that the set of features seen on the validation-function is not consistent with the ones in the training set.  For the former, it could be attributed by highly noisy feature values coming from the randomness of the first algorithm; For the latter, it is very likely that the landscape (hence the ELA features) of the test function is dissimilar to the ones in the training set. Further analyses per function/algorithm pair (Figure~\ref{fig:pred_vs_real_swpoint}) aims to investigate these two possible factors. This could in part be an artifact of the leave-one-function-out validation, since the BBOB function have been originally created such that each function has distinct high-level properties~\cite{bbobfunctions}. However, we should note that the features we consider are trajectory-based, and are thus not necessarily as different between functions as the global version of the same features would be.

Figure~\ref{fig:pred_vs_real_swpoint} show this dependence on F7 and F15. In the top subfigure (DE to PSO on F7) we see that the actual switch (blue dots) is mostly detrimental, while the predicted value is somewhat positive, which is also reflected by quite high MSE scores of the model. Note that, the relative benefit values are not considerably noisy from the chart as the majority the sample concentrates at the very bottom, which should be learnable if the RF model were trained on this function. Hence, in this case, we conclude that, in our leave-one-function-out procedure, the model fails to generalize to function F7.

In contrast, in the bottom part of Figure~\ref{fig:pred_vs_real_swpoint} (PSO to CMA on F15), we see that the overall behavior of benefit decreasing as the search continues is quite well captured by the predictions. There are two interesting aspects of the results: (1) the model seems to yield unbiased predictions of the relative benefit, which is strong support that the model generalizes well to F15; (2) The variance of the predictions are much smaller than that of the actual values, implying the possibility of a substantially large random noise when measuring the relative benefits (this observation matches with previous studies on the intrinsic large stochasticity of iterative optimization heuristics~\cite{DBLP:conf/gecco/VermettenW0DB22}). The impact of this noise might be reduced in future by performing the switch multiple times from the same switching point, leading to more stable training data.

\subsection{Impact of features}

\begin{figure}
    \centering
    \includegraphics[width=1\textwidth, trim=5mm 5mm 5mm 0mm, clip]{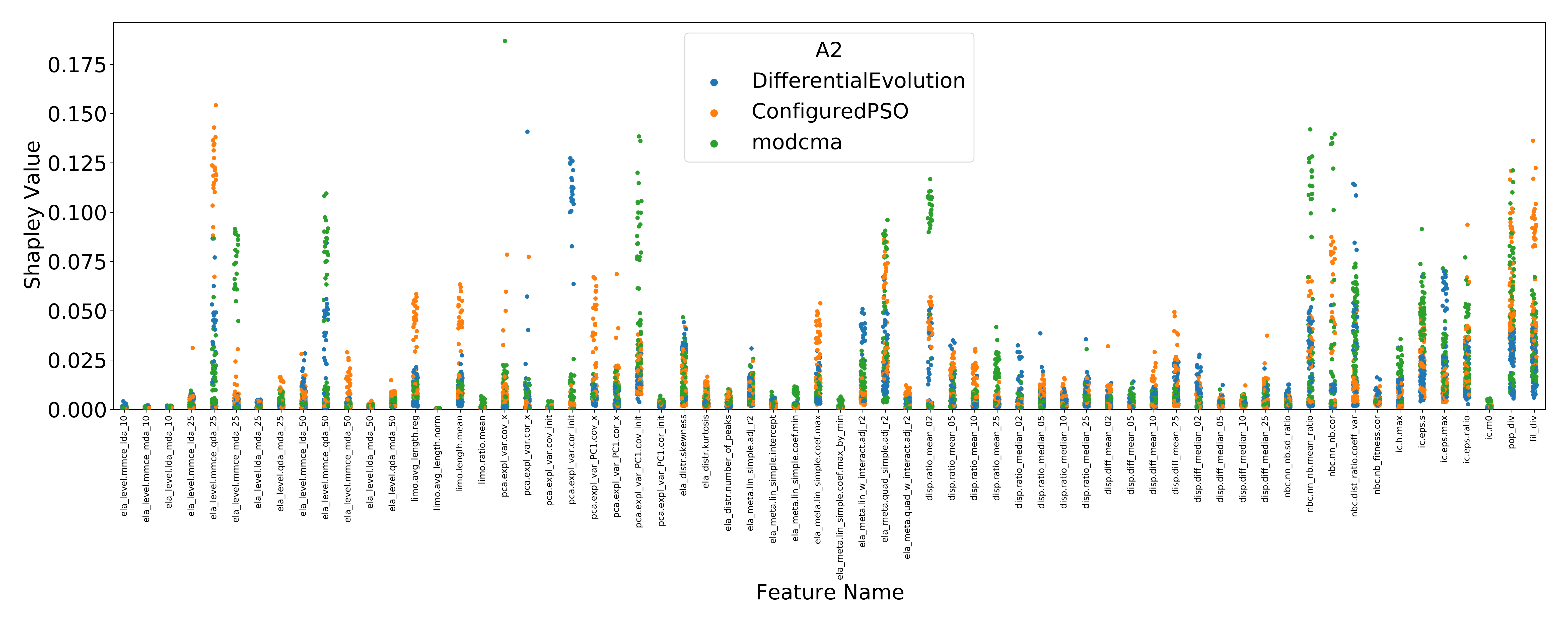}
    \caption{Shapley values of the features in each of the models which predict the real-valued improvement of the switch. Each dot corresponds to one model, trained on 23 functions, where the SHAP-values are calculated on the function which has been left out.}
    \label{fig:shaps}
\end{figure}

In addition to considering the accuracy of the trained models, we can also use the models themselves to get insights into the underlying structure of the local landscapes as seen by the algorithms. In particular, we make use of Shapley additive explanations (SHAP~\cite{NIPS2017_7062}) to gain insight into the contribution of the ELA features to the final predictions. Since we consider a multitude of models, we consider the distributions of Shapley values of each feature, aggregated across functions and algorithms. This is visualized in Figure~\ref{fig:shaps}. 

Since Figure~\ref{fig:shaps} is colored according to the algorithm being switched to, we can observe some interesting differences. Specifically, we see that the largest Shapley values are clearly present for different features depending on the $A_2$ algorithm considered. This seems to indicate that the state of the local landscape has a different effect on each algorithm. Thus, the models are indeed taking into account some specific information about the potential performance of the specific algorithm combination on which it is trained, rather than only identifying whether continuing with the current algorithm is useful in general. 

\begin{figure}
    \centering
    \includegraphics[width=\textwidth]{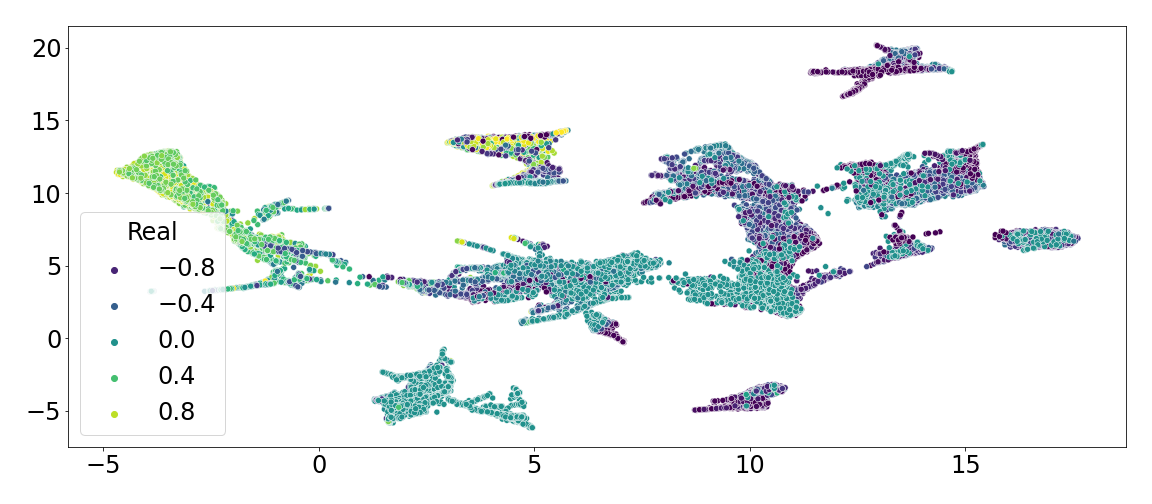}
    \caption{UMAP embedding of all datapoints from the CMA to DE model, where the color corresponds to the relative benefit of performing the switch.}
    \label{fig:umap}
\end{figure}

By considering the local landscape features themselves without taking the models into account, we can perform dimensionality reduction to judge whether there are any patterns present in the landscape which could potentially be exploited. We make use of UMAP~\cite{umap}, and visualize the features obtained during the runs of CMA-ES in Figure~\ref{fig:umap}. While this figure shows some clear clusters of similar values of the relative benefit of switching, there exist some regions where this distinction is not as clear. Based on this observation, it seems likely that the model quality can be further improved, although it is still limited by the inherent stochasticity in the dynamic algorithm selection task.

\section{Conclusions and Future Work}

We have shown that the information contained in the local landscape features collected during the run of an algorithm can be used to determine whether switching would be beneficial. This indicates that there is potential in creating algorithm switching policies based on only the local landscape seen from the perspective of the running algorithm. These policies could be constructed by defining a minimum threshold of predicted benefit when switching. While the impact of switching on the reliability of further switches remains an open question, the fact that the models are based only on samples seen instead of internal states of the algorithms seems likely to mitigate this potential issue. 

While the set of features used in this work is rather large, we observed that there are clear differences in importance to the predictions made, suggesting that a reduction of the feature space can be done without hurting the potential accuracy. This would be particularly useful when considering the dynamic algorithm selection problem from a reinforcement learning perspective, as is often the case in the configuration setting~\cite{DBLP:conf/ecai/BiedenkappBEHL20}. 

While the notion of building policies for online adaptation of algorithms seems promising, one major factor which has to be taken into account is the level of noise present. We found that the actual benefits of switching between algorithms has a very high variance, which is not perfectly captured by the created models. Further examination of the used warm-starting mechanisms and their robustness would help to make this approach more stable. 

\subsubsection{Acknowledgments}
This work was supported by a scholarship from SPECIES for Diederick Vermetten to visit Edinburgh Napier University. Parts of this work were done using the ALICE compute resources provided by Leiden University.

\bibliographystyle{splncs04.bst}
\bibliography{references_shortened.bib} 
\end{document}